\documentclass[manuscript,screen]{acmart}

\AtBeginDocument{%
  \providecommand\BibTeX{{%
    \normalfont B\kern-0.5em{\scshape i\kern-0.25em b}\kern-0.8em\TeX}}}

\setcopyright{acmcopyright}
\copyrightyear{2023}
\acmYear{2023}
\acmDOI{XXXXXXX.XXXXXX}

\acmConference[
]{
}{
}{
  }
\acmPrice{15.00}
\acmISBN{978-1-4503-XXXX-X/18/06}

\usepackage{tabularx}
\usepackage{float}
\usepackage{longtable}

\begin{document}
\title{Multi Class Depression Detection Through Tweets using Artificial Intelligence}

\author{Muhammad Osama Nusrat}
\authornote{Conceptualization: Muhammad Osama Nusrat, Waseem Shehzad; Methodology: Muhammad Osama Nusrat, Waseem Shehzad; Software: Muhammad Osama Nusrat, Saad Ahmed Jamal; Validation: Waseem Shehzad; Formal analysis: Muhammad Osama Nusrat, Waseem Shehzad; Investigation: Muhammad Osama Nusrat; Resources: Muhammad Osama Nusrat, Saad Ahmed Jamal; Data curation: Muhammad Osama Nusrat, Saad Ahmed Jamal; Visualization: Muhammad Osama Nusrat, Saad Ahmed Jamal; Supervision: Waseem Shehzad; Writing—original draft preparation: Muhammad Osama Nusrat; Writing—review and editing, Saad Ahmed Jamal.  \\ 
Declaration Statement:\\
All authors have read and agreed to the published version of the manuscript. 
}

\email{i212169@nu.edu.pk}
\orcid{0009-0007-8304-9852}
\affiliation{%
  \institution{National University of Computer and Emerging Science}
  \streetaddress{A.K Brohi Road, H 11/4}
  \city{Islamabad}
  \country{Pakistan}
  \postcode{44000}
}

\author{Waseem Shahzad}
\affiliation{%
  \institution{National University of Computer and Emerging Sciences}
  \streetaddress{A.K Brohi Road, H 11/4}
  \city{Islamabad}
  \country{Pakistan}}
\email{waseem.shahzad@nu.edu.pk}

\author{Saad Ahmed Jamal}
\affiliation{%
  \institution{University of Salzburg}
  \streetaddress{Schillerstraße 30, 5020 Salzburg}
  \city{Salzburg}
  \country{Austria}}
\email{saad.jamal@stud.plus.ac.at}
\orcid{0000-0002-4256-0298}

\begin{abstract}
 Depression is a significant issue nowadays. As per the World Health Organization (WHO), in 2023, over 280 million individuals are grappling with depression. This is a huge number; if not taken seriously, these numbers will increase rapidly. About 4.89 billion individuals are social media users. People express their feelings and emotions on platforms like Twitter, Facebook, Reddit, Instagram, etc. These platforms contain valuable information which can be used for research purposes. Considerable research has been conducted across various social media platforms. However, certain limitations persist in these endeavors. Particularly, previous studies were only focused on detecting depression and the intensity of depression in tweets. Also, there existed inaccuracies in dataset labeling. In this research work, five types of depression (Bipolar, major, psychotic, atypical, and postpartum) were predicted using tweets from the Twitter database based on lexicon labeling. Explainable AI was used to provide reasoning by highlighting the parts of tweets that represent type of depression. Bidirectional Encoder Representations from Transformers (BERT) was used for feature extraction and training. Machine learning and deep learning methodologies were used to train the model. The BERT model presented the most promising results, achieving an overall accuracy of 0.96.

\end{abstract}



\received{19 June 2023}
\received[revised]{20 November 2023}

\maketitle

\section{Introduction}

Depression is a mental disorder in which a person becomes hopeless and sad for a continuous period. Depression usually occurs due to two different scenarios that happen with the subject, and the two scenarios are quite uncommon.
The first is the reason, which can be a situation from which the subject has extracted a traumatic experience and has retained unfondly memories from which the subject cannot disconnect his or her thoughts. This situation, in turn, causes long-term retained depression. The second scenario is alienation caused by the subject’s daily life events and experiences, which in turn causes the subject to lose motivation gradually, hence causing depressing thoughts over time, and those thoughts then transform into depression in the long run. The person experiencing depression tends to have no happiness; the things that a person used to enjoy at one time and were a source of happiness for him no longer make him happy. It not only rattles a person’s mental health seriously, but it also affects the person physically, making him feel lazy; he may experience thoughts that he is useless and he lives a purposeless life. Due to depression, a person’s relationship is also damaged, and he lives in an abusive relationship. Depression could start or happen to a person if some significant incident happened, e.g., the death of someone he was close to, bad marks in exams, a sudden injury, past regrets, etc.  According to a report \cite{SproutSocial2023}, in 2023, roughly 4.89 billion people will use social media on all platforms, like Twitter and Facebook. 
This is an enormous number, and by this figure, one can imagine how important social media has become 
It is not wrong to say that it has become an integral part of our daily lives.  

The increase in the use of social media has also resulted in mental health issues \cite{RobinsonSmith2023}. Much research has been done in analyzing social media posts of users to check whether they are suffering from depression or not. In a research study, Fazida et al. \cite{karim2020social} proved that individuals who use social media have more chances of depression and anxiety. In the case of university students, when they see their assignments, homework, and other activities that cause stress, they feel overwhelmed. To relax their minds, they scroll through social media, which helps them get instant gratification and a dopamine boost. But this is temporary because as soon as one thinks again of assignment and the work one have to do, the person again rush for a dopamine boost in the form of caffeine, binge-eating, watching movies, etc., and the cycle continues. This is why most people get depression, and this is alarming.  

According to the National Vital Statistics Report, Sally et al. \cite{Curtin2020} did a study in which she aimed to compare the suicide rates of children and adults in America between the age of 10-24 years before Facebook launched from 2000 to 2017. According to the research, suicide rates were very low from 2000 to 2007, and they increased to 57\% from 2007 to 2017. According to Alexey Makarin, a professor at the Massachusetts Institute of Technology(MIT), Most students reported being mentally sick after using social media \cite{Walsh2022}. The main reason turned out to be peer comparison. Makarin and his team surveyed various universities to see Facebook's impact on students’ mental health. The researchers concluded that the number of students feeling mentally distressed and ill increased compared to last year. Moreover, the number of students who were using Facebook in college for their anxiety disorders increased to 20\%. Many students were reported to take antidepressants to cure their anxiety and mental illness.  

There are several reasons and motivations for conducting research on sentiment analysis on social media platforms, i.e. Twitter. The first reason is that more than 4.89 billion people have active social media accounts, which is huge. Traditional methods like the Diagnostic and Statistical Manual of Mental Disorders (DSM), Patient Health Questionnaire (PHQ-9), Beck Depression Inventory (BDI), Hamilton Psychiatric Rating Scale (HRSD), and clinical interviews are costly and take time. Although they are the most effective, social media sentiment analysis can also replace these effective methods in the near future. The reason why Twitter was chosen as our focus platform for depression detection is that Twitter has a large amount of open-source data that is publicly available. Secondly, data is in the form of text, is easy to handle, and to maneuver around. Moreover, several types of tweet data are available, e.g., fresh and old, which can help to draw comparisons. Twitter alone has 396.5 million 
active users, which is another reason for its selection as a platform for performing sentiment analysis. Lastly, the biggest reason is privacy. Most of the users on other platforms like Instagram and Facebook post information that is not public. The information on Twitter is public, making it a research-friendly platform.  Facebook and Instagram mostly have image-based data. Although both platforms have a more significant number of users than Twitter, there is still a drawback: they contain data in different languages, so it is a challenge to perform sentiment analysis in multiple languages. 

In this paper, the research problem was to analyze users' tweets and predict whether a person using Twitter to tweet was depressed. Also, to predict the type of depression the user was facing. A considerable amount of work has been done before to predict depression in tweets, but no work has been done to predict types of depression on Twitter. Jina et al. \cite{kim2020deep} worked on predicting the type of mental illness (anxiety. Bipolar disorder, borderline personality disorder, autism, and schizophrenia). They chose Reddit as their platform for research. Contrarily, in this research, the research problem was to predict the five most prevalent types of depression (Bipolar depression, Atypical depression, Psychotic depression, Major depressive disorder, and Postpartum depression). Moreover, it uses explainable AI to highlight parts of the tweet, due to which the model predicts the type of depression.


The first step was to scrap tweets with keywords or lexicons. Apify, which is a scrapping platform, was chosen to collect tweets. The lexicons which were used to scrap tweets were verified by the domain experts. After scrapping the tweets, preprocessing was done, which removes hashtags, URLs, @, punctuation removal, links, usernames, and stop words and converts the tweets to lowercase. The tweets which were not in English were excluded. Similarly, spammy tweets were also excluded.  After that, tweets were labeled according to the lexicons. The tweets in which the person doing the tweet is not depressed were not marked as depressed.  After labeling tweets, tokenization was done, splitting the data into train and test, and then using BERT for feature extraction and training. Finally, explainable AI was used to give reasoning if a certain tweet is marked as depressed, then what is the reason behind it? 

Following is the significant contribution to our research.

\begin{itemize}
  \item Constructing a dataset of depression along with its types from scratch.
  \item Predicting types of depression from the tweets
  \item Use of Explainable AI to Interpret Model's Decision-making Process
\end{itemize}

The research is divided into five sections. The first section discusses the introduction and background of the problem, the study's motivation, and finally, the contribution. The second section covers the literature review. A comparison table is included in which each study's strengths and weaknesses are discussed. The problem statement, research questions, and inclusion and exclusion criteria are also examined. The third section delves into research methodology, including a pipeline discussion about the dataset and the language used to execute the problem. In the fourth section, the implementation and results are explained. In the concluding section, deliberation occurs on the research's final takeaways and potential future directions.
The following research questions were addressed:

\begin{itemize}
  \item Which type of keyword or phrases are used by people suffering from a particular type of depression?
  \item Can machine learning and deep learning techniques be used to measure types of depression accurately?
  \item Can NLP accurately predict types of depression in tweets?
\end{itemize}

\section{Related Work}

Rafal et al. \cite{poswiata2022opi} participated in a competition organized by Codalab. The task was to predict the intensity of depression in the tweets, whether the person posting the tweet is severely depressed, moderately depressed, or not depressed. The event organization provided the dataset. The researchers proposed their solution in which they first fine-tuned BERT, RoBERTa, and XLNet on the provided dataset. Among all the models, the RoBERTa-large model delivered the superior results. The researchers then took one step further and then again fine-tuned RoBERTa large on the dataset provided. The Reddit mental health dataset was also used alongside the dataset provided by Codalab, the competition's organizers. The team of authors and his fellow researchers won first prize in the competition.

Safa et al. \cite{safa2022automatic} proposed a new method that can automatically collect a large number of tweets from users and then analyze them to see whether they contain depressive features.  In the tweets that were collected, the people self-reported to be depressed. A multi-modal framework was also proposed, in which they took the help of the n-gram language model and Linguistic Inquiry and Word Count (LIWC) to predict depression. For, if a user tweet contains a certain word such as feeling lonely, feeling sad, etc., it can be analyzed using the n-gram language model; similarly, by counting the frequency of words that came in a sentence using the LIWC dictionary, it can also be checked whether a post shows depressive content or not. For example, if a tweet frequently contains the words depressed, sad, or alone, there is a high probability it is depressed.

Zhao et al. \cite{zhou2021detecting} conducted research in New South Wales to analyze people's tweets to check whether people feel more depressed after COVID-19. He analyzed the tweets between Jan 2020 and May 2022. The researchers suggested a novel classification model centered around multimodal characteristics. The findings indicated an increase in depression among individuals following the COVID-19 pandemic. In this research, there was a strange trend: people remained worried even when the restrictions imposed by the government due to COVID-19 were removed. The main reason was that they were worried that the spread of COVID-19 would increase further due to relaxation. These results will help the government to know that people need mental health help and assistance.

Rissola et al. \cite{rissola2020dataset} proposed a new method that can help collect a dataset of social media posts that contain depression or not. The author emphasized that due to the lack of a dataset, there is much difficulty in building a model that can detect depression with good accuracy. Hence, the dataset shared by the author can predict depression accurately. The researchers trained their dataset using the BERT model, and the results were very good, with good accuracy, precision, recall, and F1 scores. Further research can be done using this dataset, which will also be useful for mental health doctors. This new method of automatically collecting huge datasets will assist current and upcoming researchers in building tools and apps that can predict depression accurately.

Jina Kim et al. \cite{kim2020deep} introduced a deep learning model that has the potential to determine mental health illnesses such as bipolar disorder, schizophrenia, autism, bipolar personality disorder, and autism. The author used Reddit to collect data for their research work. There are six classes in the dataset. The researchers faced a class imbalanced issue, and to cope with that issue, they used the SMOTE algorithm. They used the XG Boost classifier and CNN to classify Reddit posts. A separate binary classification model was made to classify each mental illness to improve accuracy as, in some cases, some users mention they are suffering from multiple types of mental illness, which will make the model unable to classify the data efficiently. The evaluation metrics used were accuracy, f1 score, precision, recall, etc.
Jamal et al. \cite{jamal2023data} demonstrated data fusion techniques to improve the performance of convolutional neural networks for a supervised learning problem.

Guntuku et al. \cite{guntuku2019twitter} presented a study in which it was also analyzed whether a person is depressed by looking at Facebook or Twitter profile images. The researchers chose 28749 users of Facebook who were suffering from depression and anxiety and used this dataset for training. Then, the model was validated at 887 posts of users on Twitter. The final step tested the model with 4132 distinct Twitter users. The people who posted depression-related content had their profile images with only faces of themselves. They were not seen in any group photos, and profile images were grayscale, and they had low visual harmony.  Nusrat et al. \cite{osama_unknown} fine-tuned the BERT model using a substantial dataset of text from twitter tweets that includes both textual content and emojis with the aim to predict the most relevant emoji for a given text.

Tadesse et al. \cite{tadesse2019detection} used NLP techniques to detect depression in Reddit posts. The researchers found a common term, also called a lexicon of terms, standard in Reddit posts of depressed people. The lexicon of terms was most used by depressed users in the posts. These lexicons, also called features, were found by applying machine learning and natural language techniques. The Bigram combined with SVM yielded an accuracy rate of 80\% and an F1 score of 0.8, but the ensemble model gave the highest accuracy (i.e., LIWC +LDA+ bigram) of 91\% and an f1 score of 0.93.

Islam et al. \cite{islam2018depression} used machine learning techniques to detect depression on Facebook. The people posting on Facebook sometimes express their feelings in emojis, sometimes in the form of comments, so the first step was to extract features. For that purpose, authors used LIWC to extract the features from Facebook, whether a Facebook post or people's comments. The author used machine learning techniques to classify features extracted by law. These algorithms included decision trees, SVM, KNN, and ensemble classifiers. The decision tree outperformed all other classifiers, standing as the best performer. Most of the depressive user's comments were posted from midnight to early morning (AM), while very few depressed comments were posted during daytime. Ammara et al. \cite{app10196878} used machine learning techniques for flood simulation modeling and detecting types of floods. 

Ferwerda et al. \cite{ferwerda2018you} discussed a relationship between the pictures and the personality traits of the person posting the picture. For that purpose, they focused on the Instagram platform and researched a sample of 193 Instagram profiles of users with their consent. They used Google Vision API to gather the dataset of users' images on Instagram. There were a total of 54,962 images. k mean clustering machine learning algorithm was used to do clustering means the users who posted most photos with musical instruments love adventures and new experiences. They like exploring new things. These people are willing to explore new cultures. Similarly, people who post pictures related to clothing, sports, and fitness are very self-disciplined. These people have good work ethics. People who post pictures with some electronic instrument, e.g., a picture of themselves at a concert, have a personality trait of extra-version. Similarly, people who post pictures of themselves in fashionable clothing and who participate in extracurricular activities have a personality trait of agreeableness. This means these people are cooperative and supportive. Lastly, people who have fewer pictures with clothing and more pictures with jewelry or other materialistic things have the personality trait of neuroticism, which means they are insecure low self-esteem, etc., so, in this study, the author discussed five personality traits of people by analyzing their pictures on Instagram using machine learning. 

Chen et al.\cite{chen2018tweeting} discussed that NLP techniques had been used in detecting a specific type of depression. However, only a handful of studies have employed detailed sentiment analysis techniques to identify a person's mental health through their social media posts. So, in this research, Chen and his fellow researchers used a dynamic sentiment analysis algorithm that can extract fine-grained emotions from the tweets of persons. Emotive has nine features of emotions: sad, happy, disgust, shame, surprise, fear, confusion, anger, and an overall score. So, this emotive algorithm gives fine-grained emotion scores to a tweet, indicating the tweet's dominant emotions. By fine-grained emotions, it was meant the emotions of people while writing or in their speech. Using machine learning algorithms, these emotions were used as features to diagnose people with self-reported mental health conditions. SVM and random forest classifier showed the best results.

Manoj et al. \cite{kour2022depression} discussed detecting depression from tweets using NLP. 
First, tweets were scrapped, and the necessary preprocessing steps were applied. Next, a hybrid text embedding technique was used to convert text data to numbers, including fast text + TF-IDF (Term Frequency Inverse Document Frequency). After that, classifiers based on machine learning were applied to the dataset to determine whether they contained depression. SVM and Random Forest classifiers were utilized, with the Random Forest classifier achieving the highest accuracy of 75\%.

Kumar et al. \cite{kumar2021predicting} focused on the Twitter platform to gather tweets and analyze whether they contain depression. The tweets were collected using Twitter API. These were the raw tweets; after that, they applied preprocessing techniques such as hashtag removal, mentions, and URL removal. NLTK library was used to tokenize each tweet. Frequent words like the, is, am, and are, which have no significance to determine depression, are also removed; then, the next task was to assign a lexicon score to each tweet. The tweets with the more depressive keywords had a high lexicon score. The dataset was trained using SVM and a Naive Bayes classifier, with SVM achieving an accuracy rate of 93\%.
 
Singh et al. \cite{singh2022idiap}  used ensemble learning to classify the tweets as moderately depressed, severely depressed, or not depressed. Codalab provided the dataset used for the competition. After preprocessing, the author fine-tuned BERT RoBERTa and xlnet for predicting the labels. Subsequently, the author implemented an ensemble voting classifier. Every model will predict a classification for the tweets. The label receiving the most votes will be selected, and the tweet will be labeled according to the highest vote. The accuracy was 0.6253, and the team of authors won 3rd prize.

Junyeop et al.\cite{cha2022lexicon} suggested a lexicon-oriented strategy for depression detection in Korean, English, and Japanese tweets. A lexicon means keywords or phrases people use when discussing depression in social media posts or tweets. The lexicon for depression was made in all three languages, and the psychiatrists later verified it. The data of users was collected using the Twitter API. A lexicon was employed to allocate a score to each tweet. Subsequently, the labelled dataset of tweets was used to train the machine learning model; this model is used to classify new tweets as depressed or not depressed.

Priyanka et al. \cite{arora2019mining} analyzed tweets of people suffering from depression and then applied machine learning algorithms to classify the tweets as depressed. The tweets were scrapped from Twitter, and then they were labeled. The criteria for labeling involved categorizing a tweet as "depressed" if it contained words such as "depression," "anxiety," or "mental illness," and as "not depressed" otherwise. There were a total of 3754 tweets. After doing preprocessing and feature extraction, the researchers used SVR(Support vector Regression) and multinomial naive Bayes algorithms for training and classification. Support vector regression gave the highest accuracy of 79.7\%.

Aswathy et al.\cite{aswathy2019deep} proposed an app to help detect depression. The main idea is that the user will enter several inputs into the system, such as I am feeling depressed, etc. If the sentence has depression features, the app will tell you that you have depression; it will negate that you don’t have depression. The main magic is inside the system, how it is working and how it is made. The author used a dataset of tweets, which was imbalanced 11911 tweets were normal, and 2308 tweets were depressed. An ensemble model of CNN \& LSTM was used for training the dataset of tweets posted by the user on Twitter. The ensemble model results were far better than those using a simple SVM model. The ensemble model accuracy was 0.97, far better than the SVM accuracy of 0.83.

Bata et al. \cite{bataineh2019ardep} in this study presented a new tool called Ardep, an Arabic lexicon that can identify the lexicon in Arabic that people use when they are depressed. To make Ardep, a massive dataset of tweets in the Arabic language was gathered, and five psychiatrists verified the lexicon. Ardep has 5922 lexicons that indicate depression. Ardep is very valuable and can be used to find signs of depression present in the content shared on various social media platforms of people in Arabic. It is an asset for mental health institutions in Arabic countries. 

Glen et al. \cite{coppersmith2015clpsych} participated in a CLPsych hackathon (Computational Linguistics and Clinical Psychology). The hackathon held at John Hopkins University had the objective of identifying depression and post-traumatic stress disorder (PTSD) from tweets. In easy words, the participants have to analyze the tweets of the user to see whether they are suffering from depression or not and propose a model that can give high accuracy.

Paula et al.\cite{cheng2016psychologist} discussed an app called Psychologist in a Pocket in this paper. The primary motivation behind making this app was to detect depression early. People use text messages, tweets, and Facebook posts to express their feelings. The author and his fellows gather the words that indicate depression on social media. For that purpose, they took the help of mental health physicians and students in college. The reason for this was that most depression symptoms are found in teenagers in their high schools and universities because of study pressure, insecurities, self-esteem issues, anxiety about paying loans and debts, etc., so after gathering information from students and health physicians, they were able to find a lexicon used by social media people experiencing depression and trauma. Anxiety, etc., so this application can check whether a person doing text messages or posts has symptoms of depression.
 
Ginetta et al.\cite{collo2018ketamine} discussed how threatening depression can be for humans if left untreated. It affects all parts of the human body, including a person's mood and health. Doctors prescribe medications for depression-suffering individuals, such as serotonin, but in some cases, individuals have treatment-resistant depression, and these medicines have no effect on their bodies. The percentage of people with this type of depression is 12-28\%. These patients require high-dose medication to cure illnesses, like antipsychotics and electroconvulsive therapy.
 
Choudhary et al. \cite{de2013predicting} discussed in this research about Twitter and how it can be used to predict Major depressive depression (MDD) in individuals. In this research, crowd-sourcing was used to find people suffering from depression. Crowd-sourcing means, in this context, gathering data on individuals who are depressed. The data was gathered via surveys. The researchers concluded that people experiencing depression had specific traits, which were low interaction with people and high sensitivity. These features were gathered, and then a machine learning model was trained on these attributes to predict depression if a person has these features. SVM was used, and accuracy came to be 70\%.
 
Lin et al. \cite{lin2020sensemood} proposed a depression detection app called Sense Mood, which can detect depression from textual and visual information on Twitter. Firstly, it finds and gathers textual and visual features which exhibit depression. After that, these features are combined to classify whether the individual posting the tweet is in a depressive state or not. Firstly, a dataset of tweets was gathered, i.e., the tweets which contain depression or no depression. Then, the textual and visual features were extracted using BERT and CNN. After combining the features, researchers classified the tweets using machine learning models.
 
Coppersmith et al. \cite{coppersmith2014quantifying} first showed the importance of social media and how it can be used to predict whether a person is suffering from depression. The author of this paper introduces a novel method for collecting the dataset of different types of disorders like post-traumatic stress disorder (PTSD) and seasonal affective disorder (SAD). Then, the author proposed classifiers that can be used to detect each type of depression with reasonable accuracy.
 
Victor et al. \cite{leiva2017towards} first emphasized that depression is more common in rich countries, i.e., 90\% of the well-off counties have high suicide rates in them. As previous studies have mentioned, social media is an available source of information and can be utilized to predict an individual mental state. The primary goal of the author in this paper was to diagnose depression at an early stage if an individual posts something online. If there is a sequence of depressive posts at a particular time, this will indicate that the person has a high chance of having depression. The contribution by the authors in this research was to use better feature extraction techniques, which can extract textual features of the tweets better than previous approaches, improve the accuracy of the models, and also apply a genetic algorithm and check whether it increases the classification results.
 
Akhtar et al. \cite{islam2020and} in this research did a survey or a questionnaire among university students in Bangladesh to see if they were suffering from depression or not. Four hundred seventy-six students participated in the questionnaire, and the results were shocking, as 15 percent of students suffered from moderate to high depression. The reason for the depression was the tuition fee the students had to pay during the pandemic when they suffered through the crisis.

Kecojevic et al.\cite{kecojevic2020impact} researched university students in their undergraduate degrees to see the effects of COVID-19 on them. The research was conducted on students from a college in New Jersey, as New Jersey was the most affected area by COVID-19. The survey was done among 162 students in New Jersey. In the survey, some questions were asked to the students, including their difficulties in life and their studies. Most of the students were female in the questionnaire. Students also discussed that due to online classes, their studies were affected, and they could not concentrate. Some other factors were low wages. The author emphasized that the college authority should pay attention to the students and try to resolve their mental health problems.

Burnap et al. \cite{burnap2015machine} developed a classifier that can help classify texts related to suicide and no suicide on Twitter. First, linguistic features that showed depression/suicide in the tweets were extracted. Then, the classifier was trained based on those lexical features. The classifier was able to detect suicide ideation in the tweets. The motivation of this research was to build a classifier that can be used to identify posts of individuals who have suicidal ideation to help individuals who have suicidal thoughts. This will lower the risk of suicide. 

Renata et al.\cite{rosa2016monitoring} proposed a solution that can help psychiatrists and mental health physicians detect if a person or patient has mental health issues. The person's mood can be checked by his social media posts, as most people depict and express their thoughts and feelings in their social media posts. So, if a model was devised that can first depict depression and then alert the doctors or family members of the patient about the patient's critical situation of the patient, it will help save a person's life.

Ghosh et al. \cite{ghosh2023attention}  presented a bidirectional LSTM CNN model with attention mechanisms to detect depression in the social media platform in the Bangla language. In previous studies, lexicon-based labeling was used for feature extraction. The attention mechanism is used in this research for feature extraction, as it focuses on relevant and essential parts of the text. The incorporation of attention mechanisms resulted in a notable improvement in the model's performance. The accuracy came to 96 percent.

Sooji et al. \cite{han2022hierarchical} presented a hierarchical attention mechanism for depression detection on Twitter. Previously, the main focus was to increase the classification accuracy of the machine learning classifier, but more work needs to be done on explainability. If a model is classifying a tweet to be depressed, then why is it doing so? So, in this research, the model aims to highlight parts of the tweet, looking at which it is marked as depressed or not depressed.

Zogan et al. \cite{zogan2022explainable} proposed an explainable approach called Multi-Aspect Depression Detection with Hierarchical Attention Network (MDHAN). Its purpose is to determine depression in the users' social media posts. It uses a hierarchical attention mechanism that can help find or extract essential features in the text data indicating depression. Hierarchical attention mechanisms can highlight significant words in a sentence and introductory sentences in a document. MDHAN combines demographic data, clinical data, and social media posts of the patient. After extracting features from all types of data and combining them, it gives attention scores to all the relevant parts. It highlights the features when predicting whether the input text reflects depression. In this way, mental health clinicians will be able to verify it, and they will be able to know if the model is making any mistakes.

Zucco et al. \cite{zucco2017sentiment} told the need for sentiment analysis and how it can be helpful for depression detection in people. The researchers in this paper explained the applications of sentiment analysis in depression prediction.  The researchers aimed to present a model to clinicians and mental health doctors that can tell the progress of a depression patient, whether he is recovering with the passage of time or not. Furthermore, the authors also proposed a prototype for depression detection that uses multimodal features such as facial expressions, speech, language, etc.

Liang et al. \cite{zucco2018explainable}  said sentiment analysis can be used to extract information about a user's opinion and thoughts. So, the benefit of sentiment analysis could also be taken in the field of medical science. Several deep learning algorithms have been used for sentiment analysis, but there needs to be more work done on explainability. If a model predicts an output, what is its reason? Cutting-edge deep learning models are challenging to understand, so there is a need for explainable AI models.

Bacco et al. \cite{bacco2021explainable} highlight the importance of explainable AI models and why they must be used. These models also tell us that if they made a particular decision, why did they make that decision? This will develop more trust in humans and practitioners in AI and help them figure out biases, especially in healthcare and medicine. If an app is made that can help diagnose depression in humans, there must be reasoning behind why it tells that a particular person is depressed. This will help psychiatrists to develop more trust in these systems, and in case of any mistake, they will correct it. The researchers also proposed an attention-based document classification and document summary system, which uses an attention mechanism to generate document summaries. \\ \\ \\ \\ \\ \\

\begin{longtable}{|p{1.1cm}|p{4.5cm}|p{3.5cm}|p{4cm}|}
\caption{Summary of Literature Review}  
\label{tab:CT}  
\\
\hline
\textbf{Ref No},\textbf{Year} & \textbf{Summary} & \textbf{Strengths} & \textbf{Weakness} \\
\hline
{\cite{poswiata2022opi}, 2022 } & The task was to make a model that can classify tweets as severely depressed, moderately depressed, and not depressed. The researchers fine-tuned BERT, RoBERTa, and XLnet on the dataset and then used ensemble learning to predict depression in the tweet. & The paper achieved the highest accuracy among all other papers. & Accuracy can further be improved. \\
\hline

{\cite{safa2022automatic}, 2022}    & The research aimed to identify depression in individuals based on their self-reported social media posts. 			& A multimodal framework extracted features from the textual and image data. It will capture information about features more comprehensively.			& Self-report depression diagnosis  \\
\hline

{\cite{zhou2021detecting}, 2021}    & The study aimed to investigate the dynamics of community depression in New South Wales, Australia, due to the COVID-19 pandemic.			& This study will help the government to know how much individuals are suffering from depression, and it will help the government to take steps to 			& The study was limited to only one city in Australia, which was New South Wales. This study can be extended to multiple cities.\\
\hline

{\cite{rissola2020dataset}, 2020}   & The author put forward a technique that enables the automatic collection of tweets using the Twitter API.			& The huge dataset can be used to build a robust model. 			& Twitter API gives a limited number of tweets.\\
\hline

{\cite{kim2020deep}, 2020}   & The authors introduced a deep-learning model capable of classifying different types of mental disorders (depression, anxiety, autism, mental disorders)			& The deep-learning model has the potential for utilization by mental health clinicians to detect depression in their patients.		& In this research, the authors only considered Reddit as a social media platform; moreover, the researchers did not consider the sociodemographic and regional differences.\\

\hline

{\cite{guntuku2019twitter}, 2019}   & The author presented a model that can predict depression from the user profile on Twitter.		& This is one of the few models that can predict depression using user profiles.		& The study was conducted on a low sample size of people.\\

\hline

{\cite{tadesse2019detection}, 2019}   & The researchers used Reddit as a platform to find a lexicon that depressed people use to express their emotions and then used an ensemble model of LDA, LIWC, and bigram with multilayer perceptron as a classifier.		& The ensemble model achieved higher accuracy than the individual models.	& Challenging problem not easy to implement\\

\hline

{\cite{islam2018depression}, 2018}   & The researchers used Facebook to predict depression from the Facebook posts and comments of the users. LIWC was used for feature extraction from Facebook comments and posts. Several machine learning classifiers were used, which included SVM and Decision Tree. The decision tree classifier outperformed other classifiers, &  Using multiple machine learning techniques helped the researchers determine which technique gave them good results.	&The study was conducted on a single platform. Moreover, Facebook has privacy concerns. Most of the users keep their data private.\\

\hline

{\cite{ferwerda2018you}, 2018}   & The researchers proved a correlation between users’ pictures on Instagram and their personality traits. With the consent of 193 users, their Instagram profiles were analyzed using Google Vision API. There were a total of 54,962 images which were gathered.	& The personality trait of a user can be predicted by looking at the images that they post on Instagram.	& Accuracy is low.\\

\hline

{\cite{chen2018tweeting}, 2018}   & This research aimed to identify self-reported depression and then used five machine learning classifiers to predict depression.& EMOTIVE was used as a sentiment analysis algorithm in this research, extracting fine-grained emotional features from a person’s tweets.& The data is based on self-reported statements.\\
\hline

{\cite{kour2022depression}, 2022}   & In this paper, the researcher’s main aim was to predict depression from the tweets posted by people on social media. & The model was able to predict depression from the input tweets.& Accuracy was low, i.e., 75\%.\\
\hline

{\cite{kumar2021predicting}, 2021}   & The authors collected Twitter data through the Twitter API, performed data preprocessing, extracted linguistic features, and utilized machine learning algorithms for training and evaluating prediction models. & Lexicons are also called features. The tweets were labeled depressed based on the lexicon verified by domain experts.& The study is limited to Twitter only.\\
\hline

{\cite{singh2022idiap}, 2022}   & Ensemble learning was used to classify the tweets as moderately depressed, severely depressed, or not depressed. Codalab provided the dataset used for the competition. After preprocessing, the author fine-tuned BERT RoBERTa and 
XLNET for predicting the labels. After that, the author applied an ensemble voting classifier.
 & Ensemble voting gave high accuracy rather than using individual classifiers & The dataset is small accuracy is low.\\
\hline

{\cite{cha2022lexicon}, 2022}   & Proposed a lexicon-based approach to detect depression in tweets in 
multiple languages, which were Korean, English, and Japanese.
& BERT gave the highest  F1 score & It only applies to text data.\\

\hline

{\cite{arora2019mining}, 2019}   & Analyzed tweets of people suffering from depression and then applied machine learning algorithms to classify the tweets as depressed or not. The authors employed SVM (Support Vector Machine), Multinomial Naive Bayes, and Support Vector Regression algorithms in their study. Support vector regression gave the highest accuracy of 79.7\%.
 & A novel method is discussed, which is used to classify depressed tweets from non-depressed tweets. &      The study is limited to binary classification. It can be extended to multiclass classification.\\

\hline

{\cite{aswathy2019deep}, 2019}   & used an ensemble model of CNN \& LSTM  for identifying depression
in tweets posted by the user on Twitter. The ensemble model results were far better than
those using a simple SVM model.

 & LSTM +SVM gave an accuracy of 85\%. & The quality of the dataset significantly influences the outcomes. Even ensemble models may not produce satisfactory results if the dataset is not of good quality.\\
\hline

{\cite{bataineh2019ardep}, 2019}   & A lexicon related to depression in Arabic was created so that it can be used to evaluate depression

 & Mental health physicians in Arab countries can benefit from this app. & This app can only benefit Arabic people.\\
\hline

\hline

{\cite{coppersmith2015clpsych}, 2015}   & The author participated in a hackathon called CLPsych. The objective of the hackathon was to identify depression and post-traumatic stress disorder (PTSD) by analyzing tweets. In easy words, the participants have to analyze the tweets of the user to find whether they are suffering from depression or not and propose a model that can give high accuracy.

 & The researchers collaborated, which was a positive sign. & It was only focused on predicting PTSD and depression.\\

\hline

{\cite{cheng2016psychologist}, 2016}   & The researchers proposed an app that can predict depression based on the lexicon of words it has been trained.

 & This app can be used to prevent suicide by monitoring people’s social media activities. & In the future, mEEG can also be added to the app, which will aid in detecting depression at a fast pace.\\

\hline

{\cite{collo2018ketamine}, 2018}   & The study investigates the effects of ketamine on structural plasticity in human dopaminergic neurons.
 & An important area of research has been explored, which is structural plasticity. & The results of the research cannot be generalized to everyone.\\
\hline

{\cite{de2013predicting}, 2013}   & The author discussed that Twitter could be used for predicting major depression.  Crowdsourcing was used to collect information about Twitter users who have been suffering from depression using the CESD square test.
& A statistical classifier was made that can predict that a person has depression in its initial phase and is likely to get depression if it is not controlled.
& sample size was small.\\
\hline

{\cite{lin2020sensemood}, 2020}   & The author proposed a depression detection app called Sense Mood, which can detect depression from textual and visual information on Twitter.

 &The accuracy of predicting depression is high because it extracted both textual and visual features. & Sometimes, this app will fail to distinguish between true emotions and sarcasm. \\

\hline

{\cite{coppersmith2014quantifying}, 2014}   & The author proposed a new method to collect datasets of various types of mental disorders and then classify them utilizing machine learning techniques. The data was collected from Twitter, spanning the years 2008 to 2013.

 & Authors took the help of LIWC software and statistical software, which can help them identify the common language used by individuals who are suffering from depression. It helped them find patterns that are common in the language used by people with mental health disorders people.
 
 & Most other mental disorders, like binge eating and Alzheimer’s, are rarely discussed on Twitter. Most people who have depression do not even reveal information on social media, which is also a problem.\\

\hline

{\cite{leiva2017towards}, 2017}   & The author's primary goal was to diagnose depression at an early stage of an individual posting something
online. If there is a sequence of depressive posts at a particular time, this will 
indicate that the person has a high chance of having depression.

 & Better feature extraction techniques that can extract
textual features of tweets.

 & There is still room for improving accuracy; a better classification system can be developed for the model.\\

\hline

{\cite{islam2020and}, 2020}   &  A survey was carried out among university students in Bangladesh to check whether they were suffering from anxiety and depression. If yes, why are they, and what is the reason and factors behind them?

 & Students who didn’t exercise and do their homework on time suffered from depression more than the ones who did contrary.

 & The snowball strategy was used due to less time and limited resources, rather than choosing the random sample strategy.\\
\hline

{\cite{kecojevic2020impact}, 2020}   &  A survey designed to assess the mental health of undergraduate students at New Jersey University, post-COVID.

 & This survey helped the university authority to improve students' mental health at New Jersey University.

 & It was a self-reported survey, not verified by mental health physicians whether a student is mentally upset or not. \\
 \hline

{\cite{burnap2015machine}, 2017}   &  Classify suicide-related communication on Twitter using a multi-class machine classification approach.

 & The study offers a machine learning approach to identify suicide-related communication on Twitter, which can help with the early detection and prevention of suicide.

 & The study focuses only on Twitter data, which may not represent the general population.\\
\hline

\hline

{\cite{rosa2016monitoring}, 2016}   & To develop a model that can predict whether a given text has suicidal ideations or not.

 & The authors were successful in making a classifier that can accurately distinguish whether a given text has suicide symptoms or not.

 & The research was focused on one platform only.\\

 \hline

{\cite{ghosh2023attention}, 2023}   &  The article presents a proposed model that utilizes attention mechanisms, bidirectional Long Short-Term Memory (LSTM), and Convolutional Neural Network (CNN) to detect depressive texts in Bangla language on social media platforms.

 & Proposed bidirectional LSTM and CNN model gave high accuracy on the dataset.

 & The researchers need to have more annotated Bangla text to increase accuracy. Moreover, a more diverse dataset can be made from different platforms like TikTok, Instagram, LinkedIn, Facebook, Pinterest, etc.\\

 \hline

 {\cite{han2022hierarchical}, 2022}   &  A novel language analysis method involving metaphor concept mapping to identify and analyze how individuals with depression express their emotions and experiences.

 & Enhances understanding of emotional expression in depression through metaphor analysis.

 & The researchers plan to conduct a large-scale study focusing on categorizing different characteristics of depression. To do this, they will analyze the metaphorical and cognitive expressions used by users on social media to describe their experiences with depression.\\

 \hline

{\cite{zogan2022explainable}, 2022}   & The researchers put forth a deep learning model that combines multiple aspects and features to enable the explainable detection of depression on social media platforms.

 & The model proposed in the study achieved higher accuracy than existing state-of-the-art approaches.

 & The model evaluation was conducted on a limited dataset and focused solely on detecting depression in English-language social media. As a result, its generalizability to other languages and contexts may be constrained.\\

 \hline

 {\cite{zucco2018explainable}, 2018}   &  The hybrid approach suggested in the research combines rule-based techniques and machine learning methods to enhance both the interpretability and precision of sentiment analysis, specifically in the medical field.

 &  Explainable AI has been used where the model also explains its decision.

 & The study only focuses on sentiment analysis in the medical domain and may not apply to other domains or contexts.\\

 \hline

\hline

{\cite{bacco2021explainable}, 2021}   &  The researchers proposed two transformer-based architectures for sentiment analysis classification. Additionally, they incorporated an extractive summary to provide an explanation for the model's decision-making process.

 & The model was able to achieve state-of-the-art results.

 & Computational cost is high.\\

 \hline

\end{longtable}


From table \ref{tab:CT}, 
Certain challenges can be drawn. Firstly, all the studies were focused on detecting whether a person doing a tweet is depressed or not. There was no focus on what is the type of depression the person is suffering from. Moreover, most studies do not use explainable AI to explain the depression detection process.





\subsection{Research Gap}

Previous work in the field of depression detection is based on binary classification, i.e., whether A social media post has depression or not. There are some studies in which multiclass classification is done, but it is not to classify types of depression in tweets. Most existing studies are based on classifying the severity level of depression in tweets. Our gap is to predict the five most prevalent types of depression, which are bipolar depression, atypical depression, psychotic depression, major depressive disorder, and postpartum depression, in the tweets. Similarly, very few studies have used explainable AI in their models for reasoning. This gap is also addressed in this research. Explainability was implemeneted that if the model has predicted that a certain tweet has bipolar depression 
then it will highlight the words in that tweet (i.e., these are the words due to which the model predicted this type of depression) The dataset labeling was also a crucial part as in most of the above studies if a social media post has the word depression, then that post is labeled as depressed; similarly, if a tweet has the word depressed, it is marked as depressed, which is not true in most cases. The dataset was labeled by considering the context of the whole sentence, which is also a research gap.

\subsection{Problem Statement}
Social media platforms contain tons of valuable information that can be used for sentiment analysis. Twitter was used to scrap the data in order to predict types of depression (Bipolar, Psychotic, Atypical, Postpartum, and Major depressive disorder) in the tweets. Moreover, explainable AI was used to give the reasoning for each prediction of the model.



\subsection{	Inclusion and Exclusion Criteria }
Below are the inclusion and exclusion criteria for our research.
\subsubsection{Inclusion Criteria}
\begin{itemize}
  \item The people who have self-reported being depressed currently and in the past are included in this study
  \item The tweets that contain the lexicons “I have bipolar depression,” “I am suffering from atypical depression,” and “I have a major depressive disorder” and vice versa are also included.
\end{itemize}
\subsubsection{Exclusion Criteria}
\begin{itemize}
  \item The research papers not publicly available to read are not included.
  \item 	The tweets which are not in the English language are excluded.
  \item   Spammy (which contains only hashtags) and repetitive tweets are also excluded.
   \item  Retweets are also excluded.
  \item   The incomplete tweets or tweets whose sentences are not complete are also excluded.
\end{itemize}

\section{Research Methodology}

According to the American Psychiatric Association, the term “depression” refers to a “syndrome that is characterized by a clinically significant disturbance in an individual’s cognitive abilities, emotional regulation or behavioral patterns. 
Surveys reported that 20\% of people of all ages faced some form of mental illness at some point, with approximately eight percent of adults having had severe depression. Aside from the severity of mental disorders and their impact on an individual’s psychological and physical health, social stigma or discrimination has caused individuals to be neglected by the community and to avoid taking the necessary treatments.

This is in addition to the fact that mental disorders impact an individual’s psychological and physical health, as literature has demonstrated the inherent challenges of diagnosing mental disease through social media platforms. Numerous researchers have attempted to discover crucial findings through various natural language processing methods. This presented the intrinsic difficulties of diagnosing mental problems such as depression. Acquiring adequate knowledge about the specific field of research is necessary to successfully create an accurate predictive model and extract the most prominent features in the data \cite{orabi2018deep}. Even if these features were removed, this does not guarantee that those characteristics are the primary contributors to achieving improved accuracies can be obtained. Because of these factors, this research investigates the prospect of utilizing deep neural networks because of the features learned within the design. Each step of the methodology is explained in subsequent sections.

\subsection{Dataset Collection}
Since no dataset is publicly available for the type of depression in tweets, the dataset had to be constructed from scratch. The first step was to make some lexicons that could be used to scrap the tweets. The next step was to verify the lexicons with the psychiatrist. After verification following lexicons were considered for scrapping tweets. 

Major depressive disorder: “I have a major depressive disorder”, “I am suffering from major depressive disorder”, “I have major depression,” “suffering from major depression.” “Major depressive episode”  

Bipolar disorder: “I have bipolar disorder,” “suffering from bipolar disorder,” “I have bipolar depression,” “suffering from bipolar depression,” “bipolar affective disorder,” “bipolar mood disorder,” “bipolar.”

Atypical Depression: major depression with atypical features, atypical major depression, hypersomnia, feeling sad or hopeless, increased appetite” “weight gain, feeling worthless.

Psychotic depression: psychotic depression, delusional depression, psychotic depressive disorder, melancholic depression, “I have psychosis,” “I have psychotic depression,” 

Postpartum depression: “postbirth depression,” “post-childbirth depression,” “maternal depression,” “I have postpartum depression.”

Apify is an open-source platform which was chosen for the scrapping task. One by one, each of the above lexicons was put in the search space, and the tweets were then saved in the CSV format.

\subsubsection{	Dataset Annotation}

Adding metadata or labels to a dataset to make it simpler to use and analyze within the context of machine learning and other data-driven applications is called dataset annotation. 
The scraped tweets were manually annotated in this research study. The tweets were carefully labeled, keeping in mind the context. Tweets that contained the word bipolar, atypical, or any other type of depression were not marked as depressed until and unless the context also matched the situation. Only those tweets where it is evident that a person is suffering from bipolar depression or any other kind of depression have been labeled as depressed. Therefore, the prepared data was ensured to have high quality. Therefore, it can be used for further research. A sample image of dataset annotation with depression classes is mentioned in the figure \ref{fig:Dataset1}. The remaining dataset annotation pic is attached in the appendix.

\begin{figure}[H]
    \begin{center}
        \includegraphics[width=10cm]{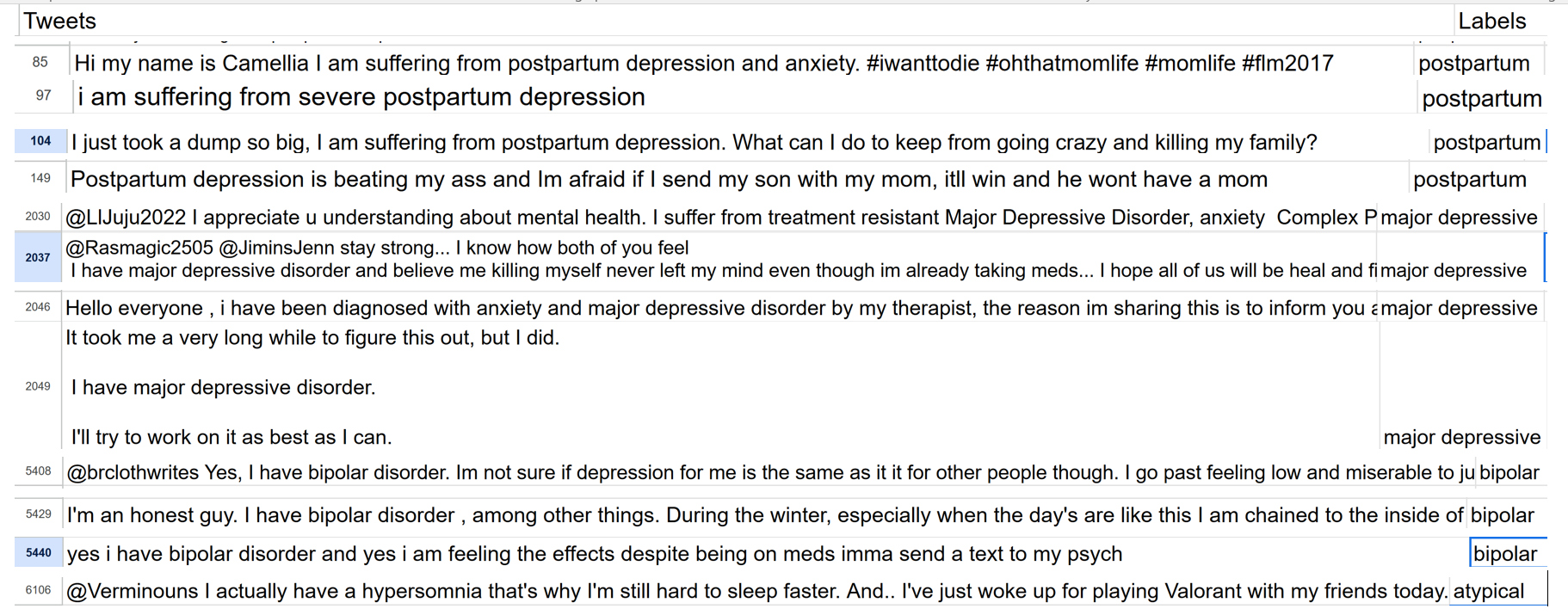}
        \caption{Dataset snippet (a)}
        \label{fig:Dataset1}
    \end{center}
\end{figure}


\subsubsection{	Data Preprocessing}
The data preprocessing phase is an integral part of depression detection through tweets. The preprocessing step involves cleaning and preparing the Twitter data for training a model. Removing noise and information useless from the tweets is also an essential part of data preprocessing. In this step, the URLs and hashtags were deleted and the tweets that contained profanity or inappropriate language. After this, the Tokenization was performed as an additional step for the data preparation. This process entails separating the text into its component subwords and providing a unique identifier for each subword. These were later split into training, validation, and testing for the deep learning model.

\subsubsection{	Features Extraction }

The process of translating raw data into valuable features that may be used for tasks involving deep learning and natural language processing is called feature extraction. Put another way, it entails selecting and extracting the essential information from the input data and displaying it appropriately in a machine-learning model. Similarly, this process can entail several strategies, depending on the kind of data and the particular natural language tasks needed. When performing tasks involving natural language processing, feature extraction may require techniques such as bag-of-words or TF-IDF to express the text data in a format that is readable by computers, or any deep learning-based model can be accomplished for feature extraction such as BERT. In this research, BERT was used for feature extraction. Natural language processing (NLP) endeavors can benefit significantly from the utilization of the sophisticated feature extractor that is BERT (Bidirectional Encoder Representations from Transformers), among other techniques such as Bag of Words (BoW), TF-IDF, and Glove. BERT is a pre-trained language model capable of learning rich and context-dependent text representations. Because of this capability, BERT is an excellent choice for feature extraction because it can be used to learn new words of text.

To use BERT as a feature extractor, the preprocessed data was fed as an input text through the pre-trained model and then acquired the contextualized embedding generated by the last layer of BERT. These embedding examples of a contextualized text representation are provided as input. Also, these embeddings that BERT caused can subsequently be used as features in downstream NLP tasks such as entity name recognition, question answering, and text classification; these embeddings were used for the type of depressions through tweets.

\subsubsection{Proposed Pipeline}

\begin{figure}[H]
    \begin{center}
        \includegraphics[width=8cm]{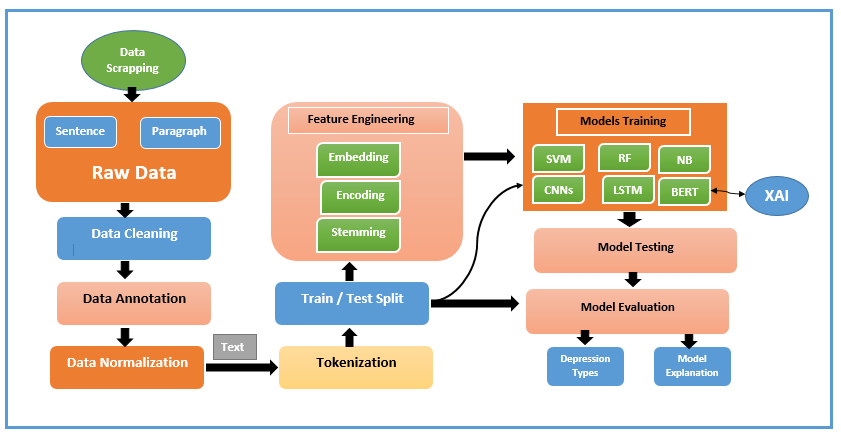}
        \caption{An overview of proposed pipeline}
        \label{fig:ProposedPipeline}
    \end{center}
\end{figure}

These NLP and machine learning-based techniques are multi-step processes that must be completed to detect the accurate depression from the tweets. The first thing that needs to be done is to compile a dataset of tweets and classify them according to whether or not they show signs of depression. This was a strenuous effort to do because depression is a complicated and varied disorder that can present itself in a variety of ways in various people. After the data has been gathered and categorized, the subsequent step was to preprocess the text and tokenize it utilizing the tokenization method. The word was broken down into individual sub-words and assigned a unique ID to each sub-word. After preprocessing, the relevant features were extracted from the text and prepared as a feature vector for the model training. Machine learning or deep learning-based language models were trained in a supervised manner. Finally, accuracy, recall, and F1 score measures evaluate the model's performance. The detail of each step of the proposed technique is explained in the subsequent section, and an overview of the proposed pipeline is mentioned in
Fig \ref{fig:ProposedPipeline}.

\subsection{Flowchart of the Proposed Solution}

\subsubsection{Data Preprocessing}

The following preprocessing techniques have been used in the dataset.

Text Normalization:
                  The tweets were converted to lowercase. Removed URLs or any web links which have no value in the text. Removed Twitter handles (e.g., @someone). Removed punctuation marks and numbers in the tweets.

Stopwords Removal:
             Removed words that are unimportant, such as the, is, etc.

\subsubsection{Data Labelling}

The tweets were labeled by reading the whole sentence context. If a tweet contains the word bipolar depression, I did not label it to be bipolar because there is a high possibility that the tweet may be about a third person who is suffering from bipolar depression, so it was highly important to read the full sentence and then label the tweet. 
The tweets were only labeled as depressed if a first-person himself/herself is suffering from depression or has suffered from depression. All the tweets which did not fall under these conditions were labeled as not depressed.
The dataset was verified by a domain expert, i.e., a psychiatrist.

\subsubsection{Feature Extraction}

BERT was used for feature extraction. Details of feature extraction are explained in Chapter 2.

\subsubsection{Tokenization}

Tokenization means to split or break the text into individual words. Tokenization is necessary because our machine learning model cannot understand long text, so text was to be converted into format that a model can process. 
Any NLP task without tokenization. In this case, BERT was used as a tokenizer. The BERT tokenizer breaks a word into a subword, which helps to overcome the problem of out-of-vocabulary words. If a word is absent in the vocabulary, it is broken down into small chunks. In this way, the BERT model will now process this word as it will have the pre-trained embedding for that chunk portion of that word. In this way Out of Vocabulary problem (OOV) is resolved.  After tokenization, padding is done to make all input text equal length, which is necessary. CLS token indicates the start of a sentence, and SEP indicates the end.
The input layer of the BERT model converts the token into a vector representation, also called embedding, which captures its meaning in the context of the sentence. A sequence of vectors in output was gained where each vector corresponds to its input token text.

\subsubsection{Attention Mechanism In BERT}
Attention is used in BERT, which calculates the weight of each input text token. 
It helps to identify which token is more important in a sentence than others. It assigns a weight or a score to each token, showing its importance. 

\subsubsection{Model Training}

The approach involved utilizing a pre-trained BERT model, which was then fine-tuned using a dataset of tweets. The BERT model was initially trained on a large corpus of text data.
Below are some of the steps for model training using BERT.

The first step is tokenization. The input tweet text is broken down into words or subwords to overcome the issue of out-of-vocabulary words. After tokenization, 
formatting was needed to be done. The BERT model takes input tokens of fixed length, so it was to be ensured that each sentence length was the same; for this purpose, padding was added to ensure this step. 
A pre-trained model of BERT was loaded and fine-tuned to the custom dataset. In other words, a pre-trained model was fine-tuned to the specific task. In this case, the task was sentiment analysis, so fine-tuning required less training data 
which was a plus point as it saved time and computational complexity.

\subsubsection{Model Evaluation and Testing}

After fine-tuning the BERT model to the dataset, the model was evaluated on the validation dataset during training. If the accuracy was not good on validation data, the hyper-parameters were adjusted until the loss was minimized. 
Finally, the fine-tuned BERT model was tested on the test dataset. Explainable AI was used later to explain why the tweets were predicted as depressed.

\subsection{Dataset and Implementation Details}
There were 14,317 tweets and six classes in the dataset.

\begin{itemize}
  \item Bipolar depression
  \item Psychotic depression
  \item Atypical depression
  \item Postpartum depression
  \item Major depressive disorder
  \item No Depression
\end{itemize}

\begin{figure}[H]
    \begin{center}
        \includegraphics[width=10cm]{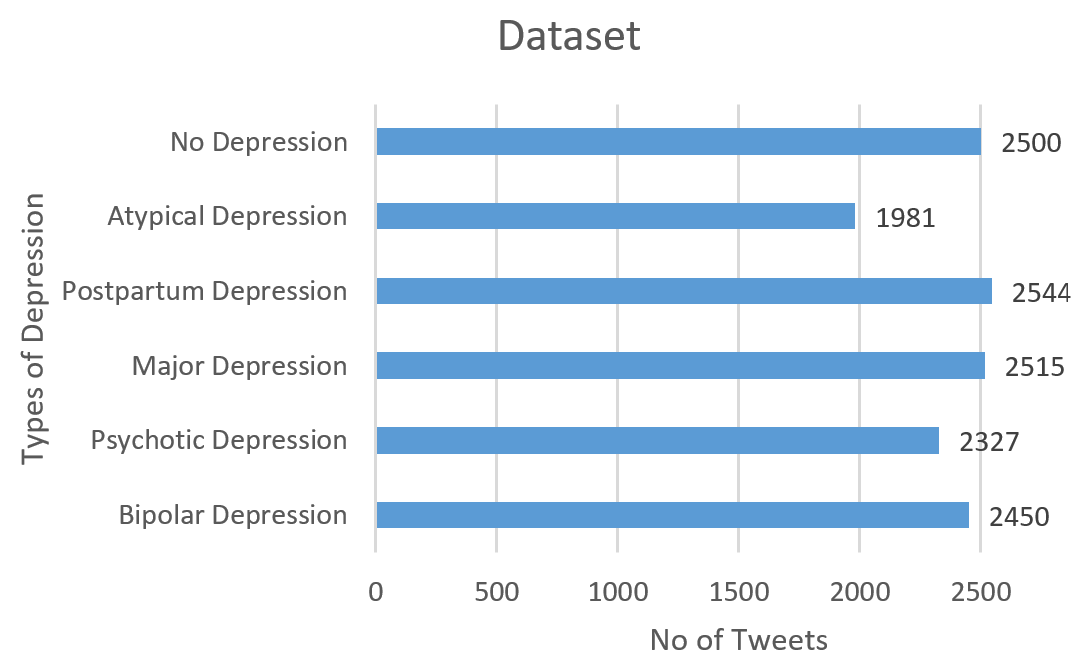}
        \caption{Dataset Details}
        \label{fig:datasetTweets}
    \end{center}
\end{figure}

Figure \ref{fig:datasetTweets} illustrates the number of tweets in each dataset class.

\section{ Experimental Evaluation}

The research aimed to evaluate the capabilities of machine learning and deep learning models for detecting depression using data collected from Twitter. The results of data analysis and model evaluation are presented in this chapter. This is followed by a discussion of the interpretation of the results and an examination of the practical implications, limitations, and potential avenues for future research. This research used 14,317 tweets from different individuals' Twitter platforms. These tweets were then subjected to preprocessing and feature engineering to extract information about depression identification. Several machine and deep learning models were used to predict depression in the dataset. This helped in determining the performance of each model on the dataset.  The performance analysis of each model was conducted using standard evaluation metrics such as precision, recall, accuracy, and F1-score. The detail of the experiments is explained in the below section concerning the individual Machine Learning and Deep learning models.

\subsection{Depression Detection Using Machine Learning}


Three Machine Learning Algorithms (Support Vector Machine, Random Forest, and Naive Bayes) were used in detecting depression on the dataset to get an estimate of which one of them gave better results.

\subsubsection{	Depression Detection Using Support Vector Machine}
Support Vector Machine (SVM) is a supervised machine learning algorithm used in various classification problems. In the case of depression detection using data from Twitter, SVM is trained using a carefully chosen subset of the dataset, specifically curated for the training process.  70\% data was used for training the model, and the rest was used for validation and test set. The performance evaluation of the model encompasses the assessment of accuracy, recall, precision, and F1-Score values. The corresponding precision, recall, F1-Score, and accuracy metrics for each type of depression are presented in the table \ref{table:EvaluatingSVMDepression}. Similarly, the overall precision, recall F1 score, and accuracy are given in table \ref{table:SVMOverall}.

\begin{table}[H]
\caption{Evaluating SVM Depression Classification Metrics}
\centering
\begin{tabular}{|p{3cm}|p{1.5cm}|p{1.5cm}|p{1.5cm}|p{1.5cm}|}
\hline
\textbf{SVM} &\textbf{Precision} & \textbf{Recall} & \textbf{F1 score} &\textbf{Support} \\
\hline
Atypical Depression&
99&
97&  
98& 
382 \\

 \hline
Bipolar Depression&
90&  
93& 
92&
484\\
\hline

Major Depression&
83&  
85& 
84&
496\\
\hline

Postpartum Depression&
91&  
88& 
90& 
503\\
\hline

Psychotic Depression&
87&  
85& 
86& 
494\\
\hline

No&
99&  
99& 
99& 
2529\\
\hline

\end{tabular}
\label{table:EvaluatingSVMDepression}
\end{table}

\begin{table}[H]
\caption{Overall Accuracy Precision Recall and F1-score of SVM}
\centering
\begin{tabular}{|p{3cm}|p{1.5cm}|p{1.5cm}|p{1.5cm}|p{1.5cm}|}
\hline
\textbf{SVM} &\textbf{Precision} & \textbf{Recall} & \textbf{F1 score} &\textbf{Accuracy} \\
\hline
Atypical Depression&
99&
97&  
98& 
382 \\
\hline

\end{tabular}
\label{table:SVMOverall}
\end{table}

\subsubsection{Depression Detection Using Random Forest}

Random Forest is recognized for its capacity to deal with high-dimensional data and intricate correlations between features. RF was trained on a portion of the dataset that was divided according to the types of depression. The performance of the Random Forest (RF) model was assessed using metrics such as Precision, Recall, F1-Score, and Support, which are presented in the table \ref{table:RandomForestEvaluating}. Table \ref{table:RandomForestOverall}  presents overall accuracy, precision, recall, and F1 score values.

\begin{table}[H]
\caption{Evaluating Random Forest Depression Classification Metrics}
\centering
\begin{tabular}{|p{3cm}|p{1.5cm}|p{1.5cm}|p{1.5cm}|p{1.5cm}|}
\hline
\textbf{Random Forest} &\textbf{Precision} & \textbf{Recall} & \textbf{F1 score} &\textbf{Support} \\
\hline
Atypical Depression&
99&
96&  
97& 
382 \\

 \hline
Bipolar Depression&
95&  
92& 
94&
484\\
\hline

Major Depression&
89&  
81& 
85&
496\\
\hline

Postpartum Depression&
91&  
90& 
90& 
503\\
\hline

Psychotic Depression&
85&  
89& 
87& 
494\\
\hline

No&
98&  
100& 
99& 
2529\\
\hline

\end{tabular}
\label{table:RandomForestEvaluating}

\end{table}
\begin{table}[H]
\caption{Overall Accuracy Precision Recall and F1-score of Random Forest Classifier}
\centering
\begin{tabular}{|p{3cm}|p{1.5cm}|p{1.5cm}|p{1.5cm}|p{1.5cm}|}
\hline
\textbf{Random Forest} &\textbf{Precision} & \textbf{Recall} & \textbf{F1 score} &\textbf{Accuracy} \\
\hline
Overall Metrics Result&
94.7&
94.7&  
94.7& 
94.7 \\
\hline

\end{tabular}

\label{table:RandomForestOverall}
\end{table}

\subsubsection{Depression Detection Using Naive Bayes}
The naive Bayes algorithm was also used for depression-type prediction in tweets. Naive Bayes first computes the prior probability for each class. In this particular case, there were six classes, so the algorithm calculates the prior probabilities based on label frequencies in the dataset. Once the prior probabilities were determined, the naive Bayes algorithm calculated the probabilities for each feature across all five depression classes. This calculation was based on understanding the patterns of how specific words are associated with particular classes. This aids in predicting the type of depression present in test tweets in the dataset. Table \ref{table:NavieBayesMetrics} shows naive Bayes depression classification metrics, and table \ref{table:NaiveBayesOverall} shows overall precision, recall f1-score, and accuracy using a naive Bayes classification algorithm.


\begin{table}[H]
\caption{Evaluating Naïve Bayes Depression Classification Metrics}
\centering
\begin{tabular}{|p{3cm}|p{1.5cm}|p{1.5cm}|p{1.5cm}|p{1.5cm}|}

\hline
\textbf{Naive Bayes} &\textbf{Precision} & \textbf{Recall} & \textbf{F1 score} &\textbf{Support} \\
\hline
Atypical Depression&
98&
91&  
94& 
382 \\

 \hline
Bipolar Depression&
94&  
77& 
85&
484\\
\hline

Major Depression&
78&  
75& 
77&
496\\
\hline

Postpartum Depression&
86&  
89& 
87& 
503\\
\hline

Psychotic Depression&
69&  
84& 
76& 
494\\
\hline

No&
98&  
98& 
98& 
2529\\
\hline

\end{tabular}

\label{table:NavieBayesMetrics}
\end{table}

\begin{table}[H]
\caption{Overall Accuracy Precision Recall and F1-score of Naïve Bayes}
\centering
\begin{tabular}{|p{3cm}|p{1.5cm}|p{1.5cm}|p{1.5cm}|p{1.5cm}|}
\hline
\textbf{Naive Bayes} &\textbf{Precision} & \textbf{Recall} & \textbf{F1 score} &\textbf{Accuracy} \\
\hline
Overall Metrics Result&
91.4&
90.9&  
91.0& 
90.9 \\
\hline

\end{tabular}

\label{table:NaiveBayesOverall}
\end{table}

\subsubsection{Machine Learning Classifiers Analysis}

Table \ref{table:MLClassifiers} gives a comparison of different machine learning classifiers which were used in this research. The algorithms' performance was assessed by measuring accuracy, precision, recall, and F1-score metrics.  For the depression detection analysis, Random Forest gave the best results. The detailed comparison of the classifier is mentioned in table \ref{table:MLClassifiers}.

\begin{table}[H]
\caption{Comparison of accuracy precision-recall \& F1 score of different machine learning classifiers}
\centering
\begin{tabular}{|p{3.2cm}|p{1.5cm}|p{1.5cm}|p{1.5cm}|p{1.5cm}|}
\hline
\textbf{Machine Learning Classifiers} &\textbf{Accuracy}&\textbf{Precision} & \textbf{Recall} & \textbf{F1 score} \\
\hline
SVM &
94.4&
94.5&  
94.4& 
94.4 \\
\hline

Random Forest&
94.7&
94.7&  
94.7& 
94.7 \\
\hline

Naive Bayes&
90.9&
91.4&  
90.9& 
91.0 \\
\hline

\end{tabular}

\label{table:MLClassifiers}
\end{table}



\subsection{Depression Detection Using Deep Learning}
 
Various deep learning models, including CNN, LSTM, and BERT, were employed to assess their performance on the dataset. Details of each of them are discussed below.

\subsubsection{Depression Detection Using Convolutional Neural Networks (CNNs)}

\begin{figure}[H]

    \begin{center}
   
        \includegraphics[width=10cm]{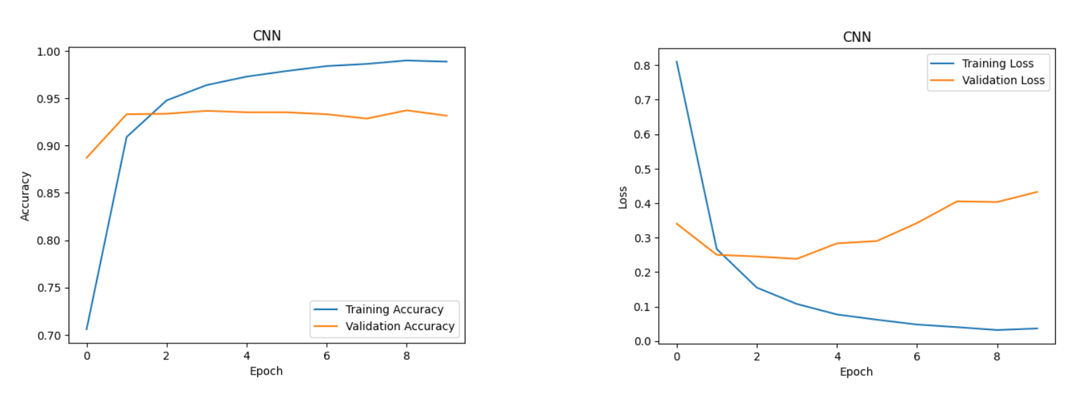}
         \caption{CNN training validation accuracy and training validation loss}
        
        \label{fig:CNNTraianingandValidationLoss}
    \end{center}
\end{figure}

The models were trained for ten epochs with thirty-two batch size.  The training process involved the utilization of cross-entropy loss and the Adam optimizer. The drop-out layer was used as the last layer, with six as a parameter to identify the depression types. The CNN model was evaluated based on the following metrics: accuracy, recall, precision, and f1-score. The training loss, validation loss, training accuracy, and validation accuracy are reported in figure \ref{fig:CNNTraianingandValidationLoss}. Table \ref{table:_CNNMetrics}  shows evaluation metric values for each type of depression, and Table \ref{table:_CNNMetrics} shows overall evaluation metric values.

\begin{table}[H]
\caption{Evaluating CNN's Depression Classification Metrics}
\centering
\begin{tabular}{|p{3cm}|p{1.5cm}|p{1.5cm}|p{1.5cm}|p{1.5cm}|}

\hline
\textbf{CNN}& \textbf{Precision} & \textbf{Recall} & \textbf{F1 score} &\textbf{Support}\\
\hline
Atypical Depression&
94.5&
94.4&  
94.4& 
382 \\

 \hline
Bipolar Depression&
94.7&  
94.7& 
94.7&
484\\
\hline

Major Depression&
91.4&  
90.9& 
91&
496\\
\hline

Postpartum Depression&
77&  
85& 
81& 
503\\
\hline

Psychotic Depression&
83&  
84& 
84& 
494\\
\hline

No&
99&  
97& 
98& 
2529\\
\hline

\end{tabular}

\label{table:_CNNMetrics}
\end{table}

\begin{table}[H]
\caption{Average Evaluation Metrics for CNN Depression Classification}
\centering
\begin{tabular}{|p{3cm}|p{1.5cm}|p{1.5cm}|p{1.5cm}|p{1.5cm}|}

\hline
\textbf{CNN}& \textbf{Precision} & \textbf{Recall} & \textbf{F1 score} &\textbf{Accuracy}\\
\hline
Overall Metrics Result&
92.9&
92.6&  
92.7& 
92.6 \\
\hline

\end{tabular}

\label{table:2_AverageCNNMetrics}
\end{table}

\subsubsection{	Depression Detection Using Convolutional Neural Networks (CNNs) with Glove}

\begin{figure}[H]
    \begin{center}
        \includegraphics[width=10cm]{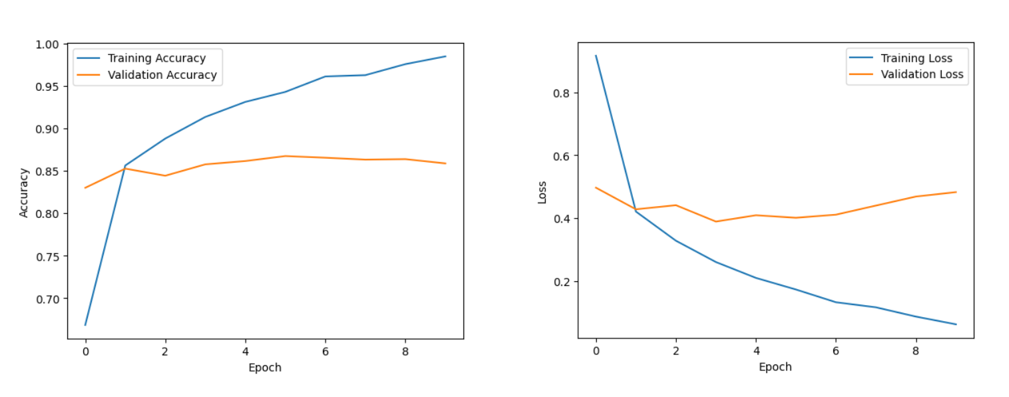}
        \caption{CNN with GloVe training validation accuracy and training validation loss}
        \label{fig:CNNGloveResult}
    \end{center}
\end{figure}

GLOVE embedding combined with CNN was employed to assess the results. The model's training configuration includes utilizing the "cross-entropy loss," the Adam optimizer, a batch size of 64, and a training period of 10 epochs. Accuracy (training and validation) and loss were illustrated in figure \ref{fig:CNNGloveResult}. Evaluation metrics for each type of depression and overall precision, recall, and f1 score accuracy are given in table \ref{fig:CNNGloveResult} and table \ref{table:_AverageEvaluationMetricsCNNGloveClassification}, respectively.


\begin{table}[H]
\caption{Evaluating CNN with Glove Depression Classification Metrics}
\centering
\begin{tabular}{|p{3cm}|p{1.5cm}|p{1.5cm}|p{1.5cm}|p{1.5cm}|}
\hline
\textbf{CNN with GloVe}& \textbf{Precision} & \textbf{Recall} & \textbf{F1 score} &\textbf{Support}\\
\hline
Atypical Depression&
86&
90&  
88& 
696 \\

 \hline
Bipolar Depression&
90&  
89& 
89&
677\\
\hline

Major Depression&
74&  
77& 
75&
613\\
\hline

Postpartum Depression&
86&  
90& 
88& 
743\\
\hline

Psychotic Depression&
90&  
74& 
81& 
474\\
\hline

No&
94&  
96& 
95& 
384
\\
\hline

\end{tabular}

\label{table:_CNNGlove}
\end{table}

\begin{table}[H]
\caption{Average Evaluation Metrics for CNN with GloVe Classification}
\centering
\begin{tabular}{|p{3cm}|p{1.5cm}|p{1.5cm}|p{1.5cm}|p{1.5cm}|}
\hline
\textbf{CNN with GloVe}& \textbf{Precision} & \textbf{Recall} & \textbf{F1 score} &\textbf{Accuracy}\\
\hline
Overall Metrics Result&
87&
86&  
86& 
86 \\
\hline

\end{tabular}

\label{table:_AverageEvaluationMetricsCNNGloveClassification}
\end{table}

Figure \ref{fig:CNNGloveResult} shows that the training and validation loss decreased as the epochs were increased. On the contrary, the accuracy (training, validation) increased with the increase in epochs.

\subsubsection{Depression Detection Using LSTM}

LSTM, short for "Long Short-Term Memory," is a recurrent neural network (RNN) used in natural language processing tasks like language modeling, machine translation, and sentiment analysis.  This study utilized two layers of the LSTM model with 64 LSTM units at each layer. The model was trained with various hyperparameters, including a dropout rate 0.2, a batch size 32, cross-entropy loss, the Adam optimizer, and ten epochs.  The evaluation metrics for each class in the dataset and overall evaluation metrics were given in table \ref{table:_LSTMClassification} and \ref{table:_Average_LSTMClassification}, respectively.

\begin{table}[H]
\caption{Average Evaluation Metrics for LSTM Classification}
\centering
\begin{tabular}{|p{3cm}|p{1.5cm}|p{1.5cm}|p{1.5cm}|p{1.5cm}|}
\hline
\textbf{LSTM}& \textbf{Precision} & \textbf{Recall} & \textbf{F1 score} &\textbf{Support}\\
\hline
Atypical Depression&
99&
97&  
98& 
382 \\

 \hline
Bipolar Depression&
93&  
90& 
92&
484\\
\hline

Major Depression&
85&  
85& 
85&
496\\
\hline

Postpartum Depression&
88&  
92& 
90& 
503\\
\hline

Psychotic Depression&
86&  
85& 
85& 
494\\
\hline

No&
99&  
99& 
99& 
2529
\\
\hline

\end{tabular}

\label{table:_LSTMClassification}
\end{table}

\begin{table}[H]
\caption{Average Evaluation Metrics for LSTM Classification}
\centering
\begin{tabular}{|p{3cm}|p{1.5cm}|p{1.5cm}|p{1.5cm}|p{1.5cm}|}
\hline
\textbf{LSTM}& \textbf{Precision} & \textbf{Recall} & \textbf{F1 score} &\textbf{Accuracy}\\
\hline
Overall Metrics Result&
94.6&
94.5&  
94.5& 
94.5 \\
\hline

\end{tabular}

\label{table:_Average_LSTMClassification}
\end{table}

\begin{figure}[H]
    \begin{center}
        \includegraphics[width=10cm]{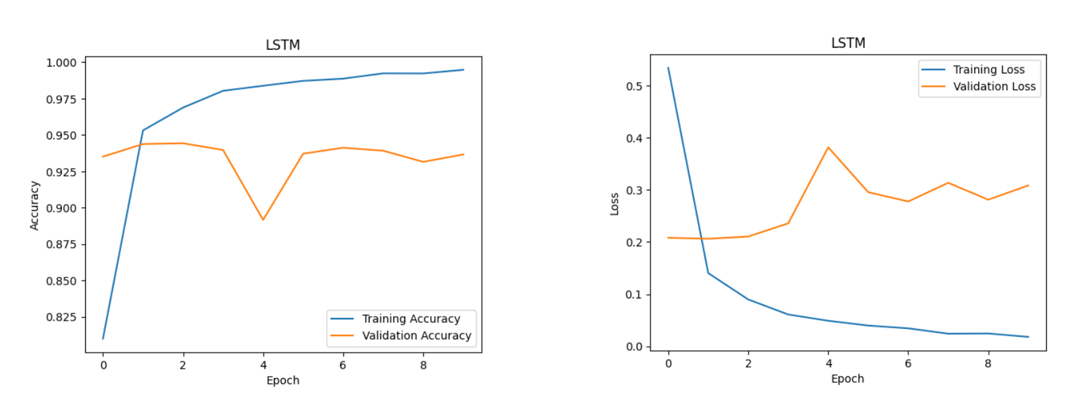}
        \caption{LSTM training validation loss}
        \label{fig:LSTMTrainingValidationLoss}
    \end{center}
\end{figure}

Figure \ref{fig:LSTMTrainingValidationLoss} shows that the training and validation loss decreases as the epochs increase. On the contrary, the accuracy (training, validation) increases with the number of epochs.

\subsubsection{Depression Detection Using LSTM with Glove}

\begin{figure}[H]
    \begin{center}
        \includegraphics[width=10cm]{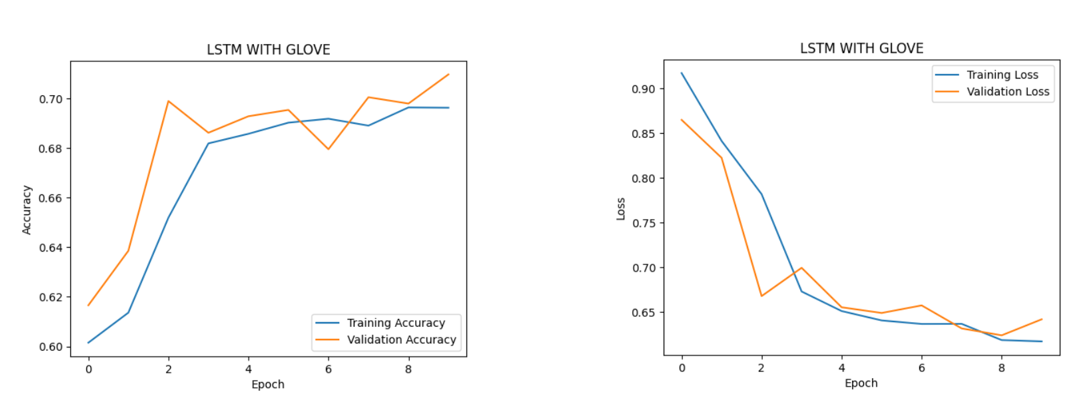}
        \caption{LSTM with GloVe training validation accuracy and training validation loss}
        \label{LSTMGloveResult}
    \end{center}
\end{figure}

GLOVE embedding combined with LSTM was used to check the accuracy score on the dataset. Using GloVe with LSTM accelerates model training and reduces computational cost. These features were then passed to fully connected layers for classification. Typically, GloVe embeddings were pretrained on extensive corpora like Wikipedia or Common Crawl and fine-tuned for specific tasks. 300 LSTM units were used with a dropout rate of 0.4. For depression classification, the dense layer was connected with the Softmax activation function. The model was trained with different hyperparameters such as “cross-entropy loss,” Adamax optimizer, 32 as batch size, and ten epochs. The evaluation metrics for each class are shown in table \ref{table:_LSTMGloveClassificationMetrics}, and for all classes, evaluation metrics are shown in table \ref{LSTMGloveResult}. 
\begin{table}[H]
\caption{Evaluating LSTM with GloVe Depression Classification Metrics
}
\centering
\begin{tabular}{|p{3cm}|p{1.5cm}|p{1.5cm}|p{1.5cm}|p{1.5cm}|}
\hline
\textbf{LSTM with GloVe}& \textbf{Precision} & \textbf{Recall} & \textbf{F1 score} &\textbf{Support}\\
\hline
Atypical Depression&
94&
96&  
95& 
382 \\

 \hline
Bipolar Depression&
0&  
0& 
0&
484\\
\hline

Major Depression&
27&  
91& 
41&
496\\
\hline

Postpartum Depression&
46&  
23& 
31& 
503\\
\hline

Psychotic Depression&
0&  
0& 
0& 
494\\
\hline

No&
98&  
99& 
99& 
2529
\\
\hline

\end{tabular}

\label{table:_LSTMGloveClassificationMetrics}
\end{table}

\begin{table}[H]
\caption{Average Evaluation Metrics for LSTM with GloVe Classification}
\centering
\begin{tabular}{|p{3cm}|p{1.5cm}|p{1.5cm}|p{1.5cm}|p{1.5cm}|}
\hline
\textbf{LSTM with GloVe}& \textbf{Precision} & \textbf{Recall} & \textbf{F1 score} &\textbf{Accuracy} \\
\hline
Overall Metrics Result &
65.5&
70.4&  
65.8& 
70.4 \\
\hline
\end{tabular}

\label{table:_LSTMGloveClassification}
\end{table}

The training validation accuracy and training and validation loss are shown in figure \ref{LSTMGloveResult}, which shows that as the epochs increase, the training and validation loss decreases. On the contrary, the accuracy (training, validation) increases with the increase in epochs.

\subsubsection{Depression Detection Using BERT}
A pre-trained BERT was fine-tuned on the new dataset. It provided the best results with an accuracy score of 0.96. This success is attributed to BERT's training on a vast corpus of text data. Furthermore, BERT effectively addresses the issue of out-of-vocabulary words.

Table \ref{table:_BERTDEpressionClassification} presents the precision, recall, F1-score, and support metrics for each type of depression. In table \ref{table:AverageEvaluationMetricsLSTMGlvoeClassification}, the combined results are displayed. BERT achieves the highest precision, recall, F1-score, and accuracy values compared to the other models utilized in the study.

\begin{table}[H]
\caption{Evaluating Bert Depression Classification Metrics
}
\centering
\begin{tabular}{|p{3cm}|p{1.5cm}|p{1.5cm}|p{1.5cm}|p{1.5cm}|}
\hline
\textbf{Bert}& \textbf{Precision} & \textbf{Recall} & \textbf{F1 score} &\textbf{Support}\\
\hline
Atypical Depression&
100&
99&  
100& 
199 \\

 \hline
Bipolar Depression&
96&  
95& 
96&
238\\
\hline

Major Depression&
87&  
88& 
88&
242\\
\hline

Postpartum Depression&
95&  
93& 
94& 
261\\
\hline

Psychotic Depression&
89&  
90& 
90& 
233\\
\hline

No&
99&  
99& 
99& 
1271
\\
\hline

\end{tabular}

\label{table:_BERTDEpressionClassification}
\end{table}

\begin{table}[H]
\caption{Average Evaluation Metrics for LSTM with GloVe Classification}
\centering
\begin{tabular}{|p{3cm}|p{1.5cm}|p{1.5cm}|p{1.5cm}|p{1.5cm}|}
\hline
\textbf{Bert}& \textbf{Precision} & \textbf{Recall} & \textbf{F1 score} &\textbf{Accuracy}\\
\hline
Overall Metrics Result&
96&
96&  
96& 
96 \\
\hline

\end{tabular}

\label{table:AverageEvaluationMetricsLSTMGlvoeClassification}
\end{table}

\subsubsection{Deep Learning Classifiers Analysis}

The detailed comparison of different deep learning classifiers which were used for predicting types of depression is mentioned in table \ref{table:Comparison of accuracy precision recall for Deep Model}.

\begin{table}[H]
\caption{Comparison of accuracy precision, recall \& F1 score of different Deep Learning Models}
\centering
\begin{tabular}{|p{3.2cm}|p{1.5cm}|p{1.5cm}|p{1.5cm}|p{1.5cm}|}
\hline
 \textbf{Deep Learning Models} &\textbf{Accuracy}& \textbf{Precision} & \textbf{Recall} & \textbf{F1 score} \\
\hline
CNN &
93&
93&  
93& 
93 \\
\hline

CNN with GloVe&
86&
87&  
86& 
86 \\
\hline

LSTM&
94&
95&  
94& 
94 \\
\hline

LSTM with GloVe &
70&
66&  
70& 
66 \\
\hline

Bert&
96&
96&  
96& 
96 \\
\hline

\end{tabular}

\label{table:Comparison of accuracy precision recall for Deep Model}
\end{table}

\subsection{Model Explainability }

Explainable Artificial Intelligence, also known as XAI, is used so that the model also tells us on what basis it made a certain decision. In other words, designing AI systems in such a way that they also explain their decision-making process to humans. In sentiment analysis, explainable AI helps to know what features or words in a sentence make a classification decision. This is done by highlighting the text in the sentence. In Python, two libraries, LIME and SHAP, are used for XAI purposes.

In the research context, XAI helps if a machine learning and deep learning model has predicted that a certain tweet contains a certain type of depression. Then, it also explains why it made that decision. It highlights the text in the tweet, saying that the model made this classification decision due to the highlighted word or words.

In figure \ref{fig:ExplainBipolar}, as shown below, the model has correctly predicted the tweet was referring to bipolar depression. The model has highlighted the words bipolar and disorder in green color, which indicates that these two words emphasize bipolar depression in the tweet. The results of the explainability of other types of depression classes are included in the appendix.

\begin{figure}[H]
    \begin{center}
        \includegraphics[width=10cm]{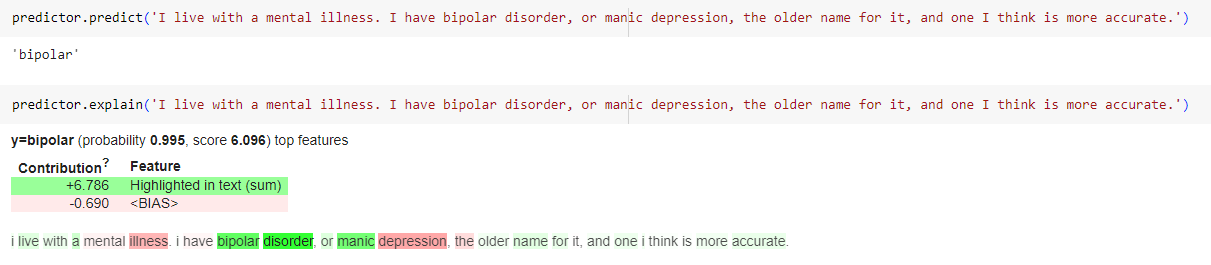}
        \caption{Explainability for bipolar depression}
        \label{fig:ExplainBipolar}
    \end{center}
\end{figure}

\section{Limitations and Future Work}

The research conducted had some limitations. Firstly, it was considered that if an individual self-reports themselves as being depressed in a tweet, that tweet is classified as a depression case right away. However, this assumption may not hold in each case. There can be moments in a person's life when one is having a bad day, but that is not necessarily linked to depression. 
Depression cannot be determined based on a single day or two. 
If an individual exhibits persistent signs of sadness over an extended duration, such as weeks or months, it may be indicative of depression. This aspect represents a limitation within the scope of this study.
There were five types of depression, which were predicted in the tweets in this research using machine learning and deep learning. The problem is that some types of depression have common features, so this was also a limitation of this research.

The dataset used was limited. For deep learning tasks, 
more data can lead to better results. In this research, as it was a novel idea, there was no existing dataset, and it had to be constructed independently. Approximately 23,000 tweets were collected, which was relatively low. In future research, results can be improved by increasing the data.

There are several other types of depression, such as Persistent Depressive Disorder (Dysthymia), Seasonal Affective Disorder (SAD), Premenstrual Dysphoric Disorder (PMDD), Subsyndromal Depression (Dysthymic Disorder NOS), Cyclothymic Depression, etc which were not addressed in this research. In the future, research can be conducted in which the above types of depression can also be predicted in tweets using Machine and Deep Learning.

\section{Conclusion}


In conclusion, both Machine Learning and Deep Learning models successfully predicted depression types based on Twitter tweets. Among the machine learning classifiers, the Random Forest Classifier yielded the highest accuracy at 94.7\%. In deep learning, BERT achieved the highest accuracy score of 96\%. The reasoning behind the model's predictions was explained using explainable AI, which further added to the usability and practicality of the models. Depression detection is a complicated and challenging disorder to diagnose. It can be challenging to accurately capture all of its facets in a single dataset. The data utilized for training depression detection models must be of high quality. It should be representative of those models sufficiently to produce accurate results. In addition, there is limited availability of labeled data, which was routed in this research by creating our own dataset. The depression database of Twitter was scrapped and made public for the research community. Still, various types of other depression are not addressed in this corpus. Other concerns of a moral and legal nature are raised in connection with the utilization of depression detection models. There are also concerns regarding the privacy and security of sensitive health data, particularly the models utilized on social media or other public datasets. These models carry with them the potential to be utilized in a manner that stigmatizes and discriminates against those who suffer from depression.

\section{Acknowledgments}

The authors would like to thank Umair Arshad at the National University of Computer and Emerging Sciences for providing us with invaluable insights and suggestions. 
Also, we would like to thank Muhammad Farrukh Bashir at Riphah International University, who helped us find students of psychology.
and Dr Ibad-ul-haq, a Consultant and a Psychiatrist, who verified the labelling of the dataset voluntarily.

\section{Funding}
There was no external funding used for this research.

\section{Code Availability}
The developed code is made available at Github: https://github.com/mnusrat786/Masters-Thesis

\bibliographystyle{ACM-Reference-Format}
\bibliography{sample-base}

\section{Appendices}

\subsection{Model Explainability For Remaining Depression Classes}

In figure \ref{fig:ExplainabilityAtypical}, as shown below, the model has correctly predicted the tweet indicates atypical depression. The model has highlighted the words hypersomnia, and if one reads the whole sentence, it was seen that the person doing the tweet is suffering from hypersomnia. Hypersomnia is a symptom of atypical depression, so this person doing the tweet is suffering from atypical depression.

\begin{figure}[H]
    \begin{center}
        \includegraphics[width=10cm]{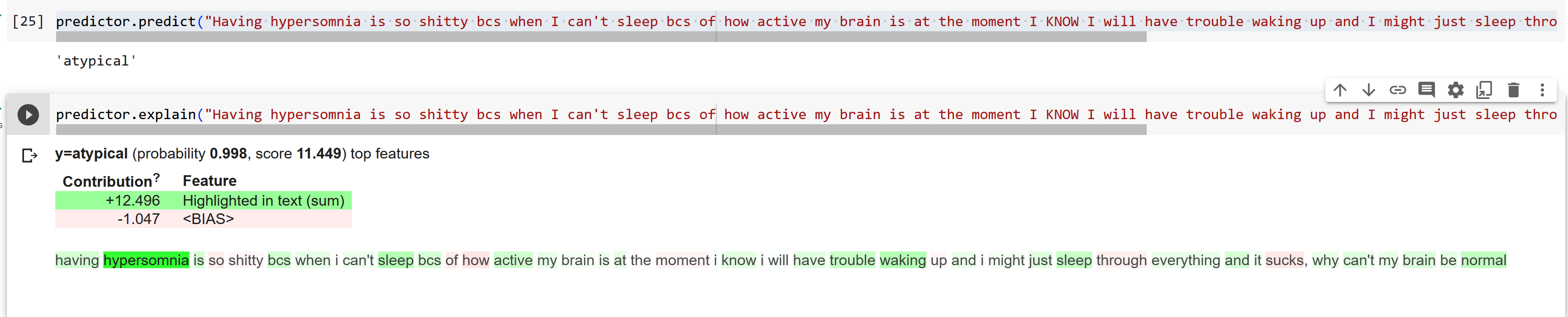}
        \caption{Explainability for atypical depression}
        \label{fig:ExplainabilityAtypical}
    \end{center}
\end{figure}

\begin{figure}[H]
    \begin{center}
        \includegraphics[width=10cm]{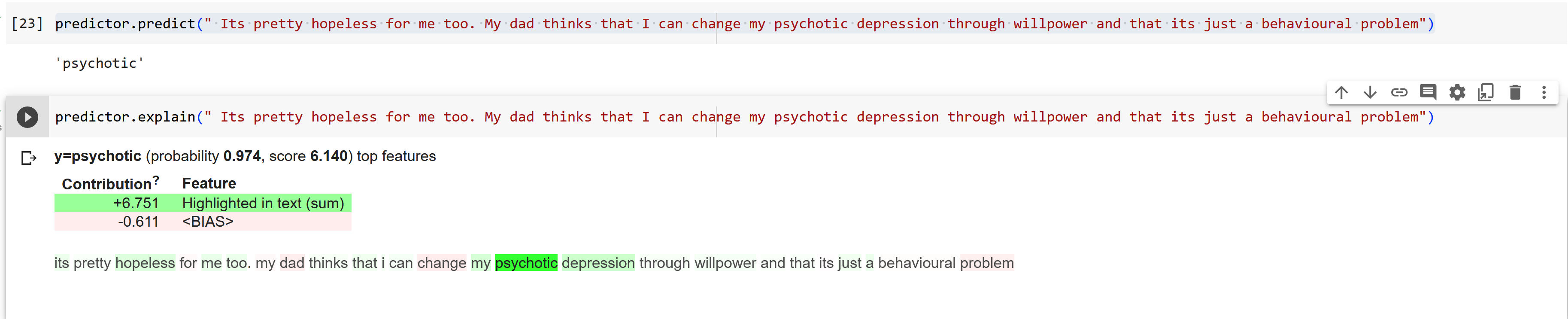}
        \caption{Explainability for psychotic depression}
        \label{fig:explainabilityPsychotic}
    \end{center}
\end{figure}

As shown in figure \ref{fig:explainabilityPsychotic}, the model has correctly predicted the tweet indicates psychotic depression. The model has highlighted the words psychotic and hopeless. If a person is feeling worthless and hopeless, then there is a high chance that the person is suffering from psychotic depression.  

\begin{figure}[H]
    \begin{center}
        \includegraphics[width=10cm]{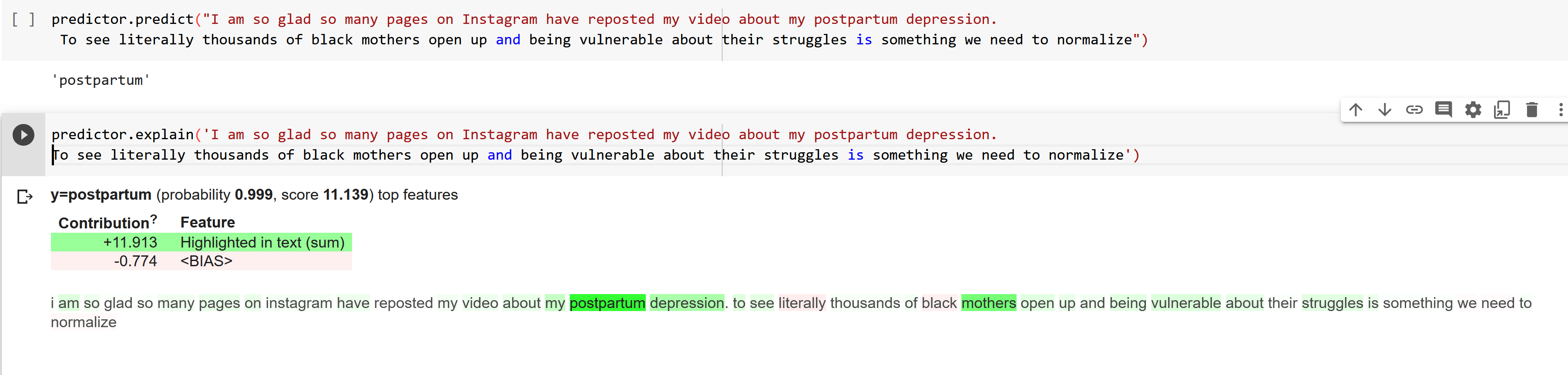}
        \caption{Explainability for postpartum depression}
        \label{fig:ExplainabilityPostpartum}
    \end{center}
\end{figure}

In figure \ref{fig:ExplainabilityPostpartum}, the model has correctly predicted that in the tweet mother is suffering from postpartum depression. Postpartum depression is mainly in mothers who are pregnant or have just given birth to their new child. The model has highlighted the words postpartum depression and mothers by looking at these words.   

In the following figure, the person doing the tweet was suffering from major depression. The model has correctly predicted it, and by using explainable AI, it has highlighted the words major depressive disorder. This means that the model made a prediction by looking at these words.
\begin{figure}[H]
    \begin{center}
        \includegraphics[width=10cm]{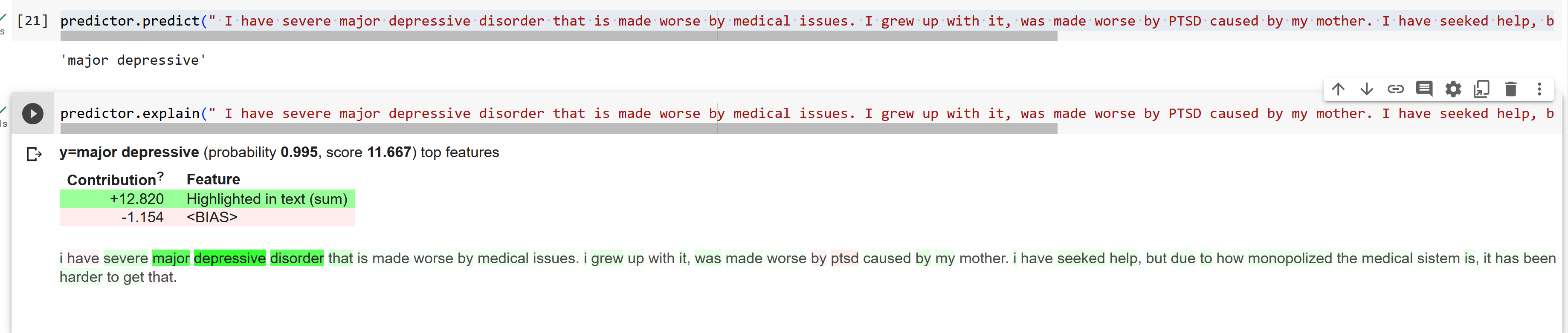}
        \caption{Explainability for major depression}
        \label{fig:ExplainabilityMajorDepression}
    \end{center}
\end{figure}
In the following figure, the model predicted the tweet has no depression. The person doing the tweet is not suffering from depression. The model decided by highlighting the words in the tweet, which are in green colour.

\begin{figure}[H]
    \begin{center}
        \includegraphics[width=10cm]{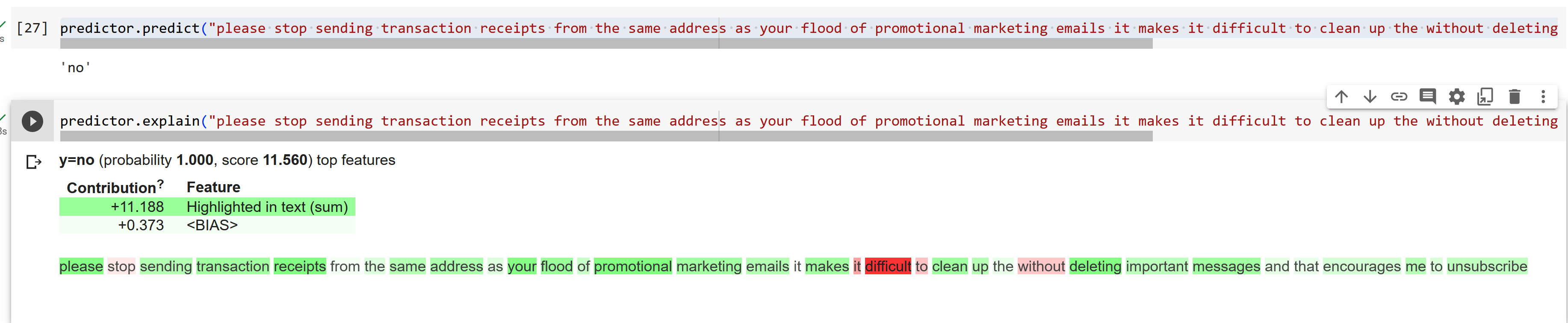}
        \caption{Explainability for no depression}
        \label{fig:ExplainabilityNoDepression}
    \end{center}
\end{figure}

\subsection{Dataset Annotation}

In the figure below, a few examples of the dataset annotation for atypical depression, psychotic depression and no depression class can be visualized.

\begin{figure}[!ht]
    \begin{center}
        \includegraphics[width=10cm]{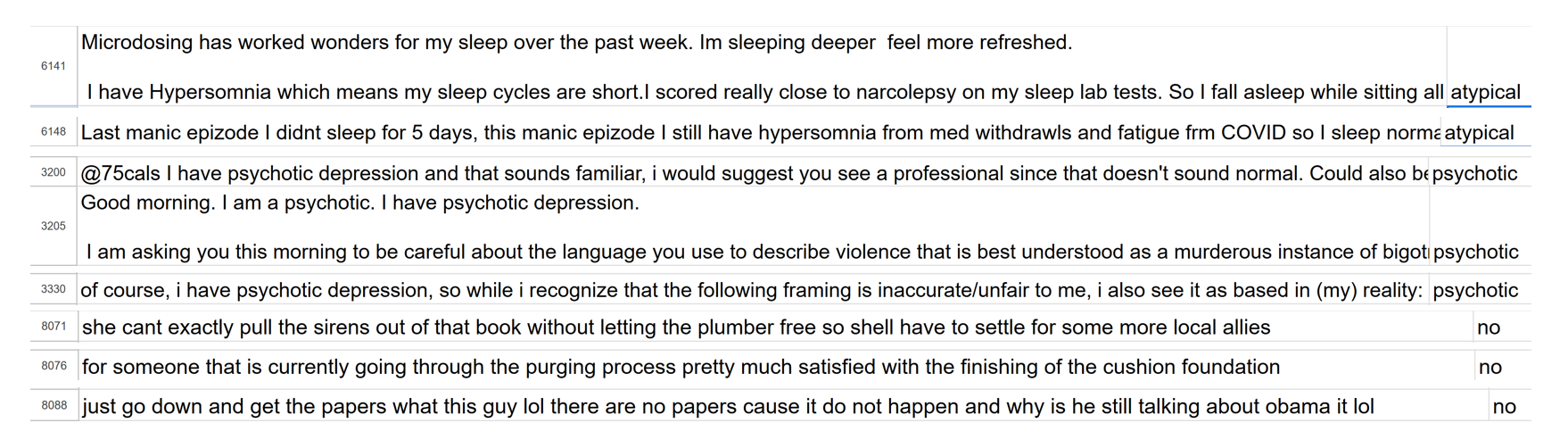}
        \caption{Dataset snippet (b)}
        \label{fig:Dataset2}
    \end{center}
\end{figure}

\end{document}